# Efficient Multiple Incremental Computation for Kernel Ridge Regression with Bayesian Uncertainty Modeling

Bo-Wei Chen*, Nik Nailah Binti Abdullah, and Sangoh Park

*Abstract*—This study presents an efficient incremental/decremental approach for big streams based on Kernel Ridge Regression (KRR), a frequently used data analysis in cloud centers. To avoid reanalyzing the whole dataset whenever sensors receive new training data, typical incremental KRR used a single-instance mechanism for updating an existing system. However, this inevitably increased redundant computational time, not to mention applicability to big streams. To this end, the proposed mechanism supports incremental/decremental processing for both single and multiple samples (i.e., batch processing). A large scale of data can be divided into batches, processed by a machine, without sacrificing the accuracy. Moreover, incremental/decremental analyses in empirical and intrinsic space are also proposed in this study to handle different types of data either with a large number of samples or high feature dimensions, whereas typical methods focused only on one type. At the end of this study, we further the proposed mechanism to statistical Kernelized Bayesian Regression, so that uncertainty modeling with incremental/decremental computation becomes applicable. Experimental results showed that computational time was significantly reduced, better than the original nonincremental design and the typical single incremental method. Furthermore, the accuracy of the proposed method remained the same as the baselines. This implied that the system enhanced efficiency without sacrificing the accuracy. These findings proved that the proposed method was appropriate for variable streaming data analysis, thereby demonstrating the effectiveness of the proposed method.

## I. Introduction

Ridge regression extends linear regression techniques, where a ridge parameter is imposed on the objective function to regularize and prevent a model [1] from overfitting. Such regularization uses $\mathcal{L}_2$ norm, or Euclidean distance, as the criterion for constraining the searching path of objective functions. Kernel Ridge Regression (KRR) further advances ridge regression by mapping feature space into hyperspace with the use of kernel functions, for example, polynomial functions and Radial Basis Functions (RBFs). In machine learning, KRR and Support Vector Machines (SVMs) have been widely used in pattern classification, especially in recent decentralized wireless sensor networks and computing platforms for the Internet of Things (IoTs).

Although KRR has a closed-form solution, which involves the inverse of matrices, calculating these inverse matrices degrades computational speeds [2]. Literature reviews [1] showed that the complexity of KRR [3] was as high as $O(N^3)$, whereas that of SVMs was $O(N^2)$, in which $N$ stands for the number of instances in data. Such a characteristic is a burden on cloud servers, which consume too much power for computation, not to mention online streaming data analysis for the IoTs [4]. The source nodes can rapidly collect information and transmit it to a fusion center, or a sink node [5, 6] (see Fig. 1), which is designed for data pooling [7]. The massive amount of streams may deplete computational resources. This requires either distributed processing [8, 9] or incremental analysis [10] to deal with big streams.

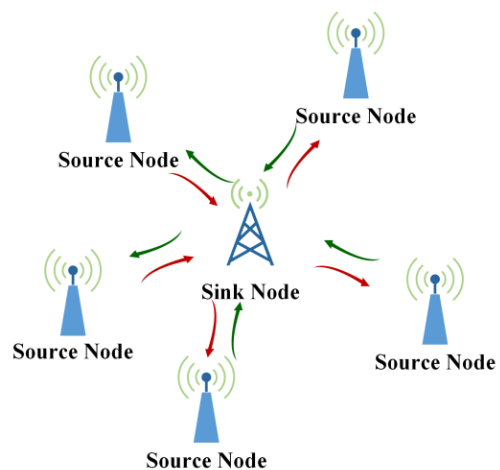

Fig. 1. Data pooling in sink nodes for wireless sensor networks

Unlike distributed processing, incremental analysis allows the system to add new training samples and to update itself without rescanning and reanalyzing existing datasets [11]. This is because incremental algorithms can reserve earlier

B.-W. Chen and N. N. B. Abdullah are with the School of Information Technology, Monash University, Australia
S. Park is with the School of Computer Engineering, Chungang University, South Korea

calculation results for updating the system when new training samples arrive in the future. To enable incremental updates, the entire mechanism involves mathematical equations with differential forms. Furthermore, based on the equations, incremental mechanisms can be classified into two types. One is single incremental (i.e., single-instance incremental), and the other is multiple incremental (i.e., multiple-instance incremental, or equivalently batch incremental). They are both conducive to relief of computational loads.

When the size of data is too large and far beyond the capability of one machine, especially when the memory space cannot accommodate the entire data at once, incremental analysis is a feasible solution. As cloud computing for the IoTs receives significant attention in recent years, more and more incremental analyses [12-21] have been devoted to this research area. Cauwenberghs and Poggio [12] established a milestone for kernelized learning as they discovered the equilibrium between existing Lagrangian multipliers and newly added ones. A differential form was derived from the cost function of SVMs and the Karush–Kuhn–Tucker (KKT) [13] conditions. Such a differential form supported single incremental and decremental learning. The derivation was shown in a subsequent study [14]. A recursive procedure was introduced to update the matrix formed by the original support vectors and the kernel matrix when a single instance was changed. The authors also devised a strategy called "bookkeeping," or the accounting strategy mentioned in [15], to determine the largest increment/decremental amount of existing Lagrangian multipliers while maintaining the equilibrium. The model by Cauwenberghs and Poggio has inspired subsequent studies, for example [14], [15], and [16]. Laskov *et al*. [15] summarized the methodology developed in [12] by presenting a systematic analytical solution. Such a solution explicitly and clearly elaborated the changes in Lagrangian multipliers with respect to three cases: Unbounded support vectors, bounded support vectors, and nonsupport vectors. Each vector was associated with one Lagrangian multiplier. Furthermore, they also presented recursive matrix updates and matrix decomposition that were conducive to incremental/decremental matrix computation. Karasuyama and Takeuchi [16] advanced the approach proposed by [12] and developed a strategy for multiple incremental/decremental learning. Multiple incremental and decremental processing were combined together during the update of the system, without being separately executed. Karasuyama and Takeuchi simplified the bookkeeping strategy mentioned in [12] by searching the shortest and easiest path when existing Lagrangian multipliers were changed. The definition of the path in their work represented a series of increment/decremental changes in the values of existing Lagrangian multipliers.

For KRR, incremental and decremental solutions become easier, when compared with those of SVMs, because KRR has a closed-form solution. Recent works, such as [17], [18], and [1], were examples for single-instance incremental regression. Based on kernel concepts, Engel *et al*. [17] developed a kernel recursive least squares algorithm, or incremental kernel regression. Their fundamental idea was equivalent to ordinary least squares (OLS) or linear least squares in statistics, but performed in hyperspace. The same algorithm was employed by Vaerenbergh *et al*. [18]. They furthered incremental kernel regression and integrated it into uncertainty analyses. However, no discussion on empirical-space or intrinsic-space computation was mentioned in [17] or [18]. In [1], a recursive version of KRR was introduced. It used a single incremental mechanism to update the weight vectors of the cost function. Moreover, a forgetting factor was integrated into the recursive form, where old and new training samples had different weights.

In frequentist methodologies, linear regression assumes there are sufficient observations. The weighting factor is calculated based on a deterministic process. Nonetheless, unlike frequentist concepts, Bayesian Regression concentrates on probabilistic modeling and Bayesian inference [19]. Given stochastic observations (i.e., predictor variables and dependent responses), Bayesian Regression examines uncertainty of a linear system by converting predictors and responses into statistical distributions. Posterior distributions derived from combination of likelihood and prior distributions are used for modeling a linear system in Bayesian Regression. As various statistical distributions can be used for modeling likelihood and prior distributions, resultant posterior distributions are different. When likelihood and prior distributions focus on Gaussian distributions, Kernelized Bayesian Regression is equivalent to Gaussian processes [20, 22]. In contrast to KRR, which is a special case of OLS, both Kernelized Bayesian Ridge Regression and Bayesian Ridge Regression are special cases of Bayesian Regression. Incremental Kernelized Bayesian Regression is computationally intensive because it has to deal with the product of a series of inverse matrices in the exponential form along with conditional means and conditional covariance matrices. Quinonero-Candela and Winther [21] proposed an incremental solution for updating the hyperparameters (i.e., means and covariance matrices) of Gaussian distributions by devising an Expectation–Maximization (EM) algorithm when marginal likelihood distributions were computed. In this study, a multiple incremental mechanism is plugged into the updating process of Incremental Kernelized Bayesian Regression.

The contributions of this study are listed as follows.
- ■ The proposed method supports both incremental and decremental analyses for multiple samples. A large dataset can be divided into subsets and fed into the system batch by batch. This enhances performance.
- ■ The proper size of a batch for incremental and decremental learning in intrinsic and empirical space is derived in this article.
- ■ Multiple incremental and decremental analyses are integrated together to update the system at the same time. Decremental learning becomes necessary when removal of unnecessary outliers is performed.
- ■ The proposed mechanism furthers the earlier version of incremental Gaussian processes by introducing incremental/decremental mechanisms and batch learning. This speeds up the uncertainty computation.

The rest of this paper is organized as follows. Section II introduces the multiple incremental/decremental computation in intrinsic space for KRR, whereas Section III then describes the details of computation in empirical space. Next, Section IV extends the proposed mechanism to Kernelized Bayesian Regression. Section V shows experimental results. Conclusions are finally drawn in Section VI.

## II. INCREMENTAL/DECREMENTAL KERNEL RIDGE REGRESSION IN INTRINSIC SPACE

KRR has two types of operation modes. One is in intrinsic space, and the other is in empirical space. Intrinsic space is used to describe dispersion matrices, also called intrinsic covariance matrices, computed based on the intrinsic dimensions of samples [1, 23]. In contrast, empirical space refers to dispersion matrices, or empirical covariance matrices, computed based on the number of samples [1, 24].

In KRR, after feature mapping by using kernel functions $\phi$, intrinsic-space computation yields favorable complexity if the number of data $N$ is far larger than the feature dimension $M$. Otherwise, empirical-space operations should be used.

Let $\{(\mathbf{x}_i, y_i)| i = 1,\ldots,N\}$ denote a pair of an $M$-dimensional feature vector $\mathbf{x}_i$ and its corresponding label $y_i$, where $i$ specifies the indices of $N$ training samples. The objective of KRR is to minimize the following cost function of least squares errors (LSEs).

$$\min_{\mathbf{u},b} E_{\text{KRR}}(\mathbf{u},b) = \min_{\mathbf{u},b} \left\{ \sum_{i=1}^{N} \left( \mathbf{u}^T \phi(\mathbf{x}_i) + b - y_i \right)^2 + \rho \|\mathbf{u}\|^2 \right\} \quad (1)$$

where $E_{\text{KRR}}$ is the cost function, $\mathbf{u}$ represents a $J$-by-1 weight vector, $\phi(\mathbf{x}_i)$ denotes the intrinsic-space feature vector of $\mathbf{x}_i$, $b$ is a bias term, and $\rho$ specifies the ridge parameter. Besides, T means the conjugate operator, and $\|\cdot\|^2$ calculates $\mathcal{L}_2$ norm. Notably, $J$ is the degree of intrinsic space when feature vectors are transformed by a kernel function.

Equation (1) can be rewritten as a matrix form, i.e.,

$$E_{\text{KRR}}(\mathbf{u},b) = \left\| \mathbf{\Phi}^T \mathbf{u} + b\mathbf{e}^T - \mathbf{y}^T \right\|^2 + \rho \|\mathbf{u}\|^2 \quad (2)$$

where $\mathbf{e}$ is a row vector of all ones. Individually differentiating (16) with respect to $\mathbf{u}$ and $b$ followed by zeroing both equations gives

$$\mathbf{u} = \left( \mathbf{\Phi}\mathbf{\Phi}^T + \rho \mathbf{I} \right)^{-1} \mathbf{\Phi} \left( \mathbf{y}^T - b\mathbf{e}^T \right) \quad (3)$$

and

$$b = \frac{1}{N} \left( \mathbf{e}\mathbf{y}^T - \mathbf{e}\mathbf{\Phi}^T \mathbf{u} \right). \quad (4)$$

Notice that $\mathbf{K} = \mathbf{\Phi}^T\mathbf{\Phi}$ instead of $\mathbf{\Phi}\mathbf{\Phi}^T$ mentioned in (3). Unlike the solution to kernel regression, i.e., $\mathbf{u} = (\mathbf{\Phi}\mathbf{\Phi}^T)^{-1}\mathbf{\Phi}(\mathbf{y}^T - b\mathbf{e}^T)$ where $\mathbf{\Phi}\mathbf{\Phi}^T$ could be singular, KRR avoids such a problem by adding a ridge term inside. The solution to (3) and (4) can be obtained by solving a system of linear equations.

$$\begin{bmatrix} \mathbf{u} \\ b \end{bmatrix} = \begin{bmatrix} \mathbf{S} & \mathbf{\Phi}\mathbf{e}^T \\ \mathbf{e}\mathbf{\Phi}^T & N \end{bmatrix}^{-1} \begin{bmatrix} \mathbf{\Phi}\mathbf{y}^T \\ \mathbf{e}\mathbf{y}^T \end{bmatrix} \quad (5)$$

where $\mathbf{S}$ denotes $\mathbf{\Phi}\mathbf{\Phi}^T + \rho\mathbf{I}$.

Based on the Schur complement theory,

$$\begin{bmatrix} \mathbf{S} & \mathbf{\Phi}\mathbf{e}^T \\ \mathbf{e}\mathbf{\Phi}^T & N \end{bmatrix}^{-1} = \begin{bmatrix} \mathbb{A} & \mathbb{U} \\ \mathbb{V} & \mathbb{N} \end{bmatrix}^{-1}$$

$$= \begin{bmatrix} \mathbb{M} & -\mathbb{M}\mathbb{U}\mathbb{N}^{-1} \\ -\mathbb{N}^{-1}\mathbb{V}\mathbb{M} & \mathbb{N}^{-1}\mathbb{V}\mathbb{M}\mathbb{U}\mathbb{N}^{-1} + \mathbb{N}^{-1} \end{bmatrix} \quad (6)$$

where

$$\begin{aligned} \mathbb{M} &= \left( \mathbb{A} - \mathbb{U}\mathbb{N}^{-1}\mathbb{V} \right)^{-1} \\ &= \mathbb{A}^{-1} + \mathbb{A}^{-1}\mathbb{U}\left( \mathbb{N} - \mathbb{V}\mathbb{A}^{-1}\mathbb{U} \right)^{-1}\mathbb{V}\mathbb{A}^{-1} \\ &= \mathbf{S}^{-1} + \mathbf{S}^{-1}\mathbf{\Phi}\mathbf{e}^T \left( \mathbf{e}\mathbf{e}^T - \mathbf{e}\mathbf{\Phi}^T\mathbf{S}^{-1}\mathbf{\Phi}\mathbf{e}^T \right)^{-1} \mathbf{e}\mathbf{\Phi}^T\mathbf{S}^{-1} \end{aligned} \quad (7)$$

This form becomes useful in the following incremental and decremental processes as $\mathbf{S}^{-1}$ is repeatedly used in the process rather than $\mathbf{S}$.

### A. Single Incremental and Decremental Processes

For intrinsic space, single incremental and decremental processes are straightforward. During the incremental phase, given a new training sample $(\mathbf{x}_c, y_c)$, the update of (3) becomes

$$\begin{aligned} \mathbf{u}[\ell+1] = \\ \mathbf{S}[\ell+1]^{-1} \mathbf{\Phi}[\ell+1] \left( \mathbf{y}^T[\ell+1] - b[\ell+1]\mathbf{e}^T[\ell+1] \right) \end{aligned} \quad (8)$$

and

$$\begin{aligned} b[\ell+1] = \\ \frac{1}{N+1} \left( \mathbf{e}[\ell+1]\mathbf{y}^T[\ell+1] - \mathbf{e}[\ell+1]\mathbf{\Phi}^T[\ell+1]\mathbf{u}[\ell+1] \right) \end{aligned} \quad (9)$$

where

$$\begin{cases} \mathbf{S}[\ell+1]^{-1} = \left(\mathbf{S}[\ell] + \phi(\mathbf{x}_c)\phi(\mathbf{x}_c)^T\right)^{-1} \\ \mathbf{\Phi}[\ell+1] = \left[\mathbf{\Phi}[\ell] \quad \phi(\mathbf{x}_c)\right] \\ \mathbf{y}[\ell+1] = \left[\mathbf{y}[\ell] \quad y_c\right] \end{cases} \quad (10)$$

In (8)–(10), $\ell$ denotes the current state of the system, and $\ell+1$ is the next state. To save computation of $\mathbf{S}^{-1}$, the Sherman-Morrison formula and Woodbury matrix identity [25] indicate that

$$\begin{aligned} \mathbf{S}[\ell+1]^{-1} &= \left(\mathbf{S}[\ell] + \phi(\mathbf{x}_c)\phi(\mathbf{x}_c)^T\right)^{-1} \\ &= \mathbf{S}[\ell]^{-1} - \frac{\mathbf{S}[\ell]^{-1}\phi(\mathbf{x}_c)\phi(\mathbf{x}_c)^T\mathbf{S}[\ell]^{-1}}{1 + \phi(\mathbf{x}_c)^T\mathbf{S}[\ell]^{-1}\phi(\mathbf{x}_c)} \end{aligned} \quad (11)$$

Regarding the decremental phase, given an index $r$ of a sample, where $r \in \{1,\ldots,N\}$, a recursive form is created by considering the $r$th sample:

$$\begin{aligned} \mathbf{S}[\ell-1]^{-1} &= \left(\mathbf{S}[\ell] - \phi(\mathbf{x}_r)\phi(\mathbf{x}_r)^T\right)^{-1} \\ &= \mathbf{S}[\ell]^{-1} + \frac{\mathbf{S}[\ell]^{-1}\phi(\mathbf{x}_r)\phi(\mathbf{x}_r)^T\mathbf{S}[\ell]^{-1}}{1 - \phi(\mathbf{x}_r)^T\mathbf{S}[\ell]^{-1}\phi(\mathbf{x}_r)} \end{aligned} \quad (12)$$

For $\mathbf{\Phi}[\ell-1]$ and $\mathbf{y}[\ell-1]$, we simply remove the corresponding column and row from $\mathbf{\Phi}[\ell]$ and $\mathbf{y}[\ell]$, respectively.

### B. Multiple Incremental and Decremental Processes

For multiple incremental and decremental processes, assume the system is about to add $|C|$ new samples and remove $|R|$ existing data. The operator $|\cdot|$ denotes the size. Additionally, $C$ and $R$ are the sets that contain sample indices. Then, (11) and (12) respectively become

$$\begin{aligned} \mathbf{S}[\ell+1]^{-1} &= \left(\mathbf{S}[\ell] + \mathbf{\Phi}_C\mathbf{\Phi}_C^T\right)^{-1} \\ &= \mathbf{S}[\ell]^{-1} - \mathbf{S}[\ell]^{-1}\mathbf{\Phi}_C\left(\mathbf{I} + \mathbf{\Phi}_C^T\mathbf{S}[\ell]^{-1}\mathbf{\Phi}_C\right)^{-1}\mathbf{\Phi}_C^T\mathbf{S}[\ell]^{-1} \end{aligned} \quad (13)$$

and

$$\begin{aligned} \mathbf{S}[\ell-1]^{-1} &= \left(\mathbf{S}[\ell] - \mathbf{\Phi}_R\mathbf{\Phi}_R^T\right)^{-1} \\ &= \mathbf{S}[\ell]^{-1} + \mathbf{S}[\ell]^{-1}\mathbf{\Phi}_R\left(\mathbf{I} - \mathbf{\Phi}_R^T\mathbf{S}[\ell]^{-1}\mathbf{\Phi}_R\right)^{-1}\mathbf{\Phi}_R^T\mathbf{S}[\ell]^{-1} \end{aligned} \quad (14)$$

To facilitate multiple incremental and decremental processes at once, combination of (13) and (14) is necessary. Let $\mathbf{\Phi}_\mathcal{H} = [\mathbf{\Phi}_C \mid \mathbf{\Phi}_R]$ represent the concatenation of all the column vectors in $\mathbf{\Phi}_C$ and $\mathbf{\Phi}_R$. Also denote $\mathbf{\Phi}'_\mathcal{H} = [\mathbf{\Phi}_C \mid -\mathbf{\Phi}_R]^T$ as the concatenation of all the column vectors in $\mathbf{\Phi}_C$ and $-\mathbf{\Phi}_R$. Therefore, combination of (13) and (14) becomes

$$\begin{aligned} &\mathbf{S}[\ell+1]^{-1} \\ &= \left(\mathbf{S}[\ell] + \mathbf{\Phi}_C\mathbf{\Phi}_C^T - \mathbf{\Phi}_R\mathbf{\Phi}_R^T\right)^{-1} \\ &= \left(\mathbf{S}[\ell] + \mathbf{\Phi}_\mathcal{H}\mathbf{\Phi}'_\mathcal{H}\right)^{-1} \\ &= \mathbf{S}[\ell]^{-1} - \mathbf{S}[\ell]^{-1}\mathbf{\Phi}_\mathcal{H}\left(\mathbf{I} + \mathbf{\Phi}'_\mathcal{H}\mathbf{S}[\ell]^{-1}\mathbf{\Phi}_\mathcal{H}\right)^{-1}\mathbf{\Phi}'_\mathcal{H}\mathbf{S}[\ell]^{-1} \end{aligned} \quad (15)$$

For $\mathbf{\Phi}[\ell+1]$ and $\mathbf{y}[\ell+1]$, the system can simply remove the corresponding column(s) and row(s) from $\mathbf{\Phi}[\ell]$ and $\mathbf{y}[\ell]$, respectively. Subsequently, new samples are appended to the end of $\mathbf{\Phi}[\ell]$ and $\mathbf{y}[\ell]$ to generate $\mathbf{\Phi}[\ell+1]$ and $\mathbf{y}[\ell+1]$.

The batch sizes of $\mathbf{\Phi}_C$ and $\mathbf{\Phi}_R$, i.e., $|C|$ and $|R|$, can be different. Notably, the left-hand side of the two equations in (13) and (14) needs $O(J^3)$, whereas the inverse on the right-hand side requires $O(|C|^3)$ for (13) and $O(|R|^3)$ for (14), respectively [26]. To ensure performance, when the number of samples in a batch is smaller than the size of intrinsic-space features (i.e., $|C| < J$ and $|R| < J$), the system should perform an update if incremental and decremental computation is separate. For (15), $|\mathcal{H}|$ should be smaller than $J$. This implies a suitable batch size for time-series data, where new samples are rapidly generated and accumulated.

### III. INCREMENTAL/DECREMENTAL KERNEL RIDGE REGRESSION IN EMPIRICAL SPACE

According to the Learning Subspace Property in [1], the weight vector $\mathbf{u}$ has the following relation between $\mathbf{\Phi}$ and an unknown $N$-dimensional vector $\mathbf{a}$.

$$\mathbf{u} = \mathbf{\Phi}\mathbf{a}. \quad (16)$$

Combining (2) and (16) yields

$$E'_{\text{KRR}}(\mathbf{a},b) = \left\|\mathbf{K}\mathbf{a} + b\mathbf{e}^T - \mathbf{y}^T\right\|^2 + \rho\mathbf{a}^T\mathbf{K}\mathbf{a}. \quad (17)$$

Rearranging the equations after differentiating (17) with respect to $\mathbf{a}$ and $b$ yields

$$\mathbf{a} = (\mathbf{K} + \rho\mathbf{I})^{-1}\left(\mathbf{y}^T - b\mathbf{e}^T\right) \quad (18)$$

and

$$b = \frac{\mathbf{y}(\mathbf{K}+\rho\mathbf{I})^{-1}\mathbf{e}^{\mathrm{T}}}{\mathbf{e}(\mathbf{K}+\rho\mathbf{I})^{-1}\mathbf{e}^{\mathrm{T}}}. \quad (19)$$

### A. Single Incremental and Decremental Processes

Given a new training sample $(\mathbf{x}_c, y_c)$, the incremental phase is listed as follows.

$$(\mathbf{K}[\ell+1]+\rho\mathbf{I}[\ell+1])^{-1} = \begin{bmatrix} \mathbf{K}[\ell]+\rho\mathbf{I}[\ell] & \boldsymbol{\eta}_{:,c} \\ \boldsymbol{\eta}_{:,c}^{\mathrm{T}} & K_{c,c}+\rho \end{bmatrix}^{-1} \quad (20)$$

where ":" signifies all the training samples except the new one, and $\boldsymbol{\eta}_{:,c}$ is part of the kernel matrix only based on the new sample. For simplicity, let $\mathbf{Q}$ denote $\mathbf{K}+\rho\mathbf{I}$ and $Q_{c,c}$ represent $K_{c,c}+\rho$. Then, (20) becomes

$$\mathbf{Q}^{-1}[\ell+1] = \begin{bmatrix} \mathbf{Q}[\ell] & \boldsymbol{\eta}_{:,c} \\ \boldsymbol{\eta}_{:,c}^{\mathrm{T}} & Q_{c,c} \end{bmatrix}^{-1}. \quad (21)$$

However, (20) does not save computational loads as the system calculates the inverse again. According to the Sherman-Morrison formula and Woodbury matrix identity [15, 25], the inverse in (21) can be decomposed to two states. One is the current state $\mathbf{Q}^{-1}[\ell]$, and the other is $\mathbf{Q}^{-1}[\ell+1]$, shown as follows.

$$\begin{aligned}\mathbf{Q}^{-1}[\ell+1] &= \begin{bmatrix} \mathbf{Q}[\ell] & \boldsymbol{\eta}_{:,c} \\ \boldsymbol{\eta}_{:,c}^{\mathrm{T}} & Q_{c,c} \end{bmatrix}^{-1} \\ &= \begin{bmatrix} \mathbf{Q}^{-1}[\ell]+z^{-1}\mathbf{G}_{:,c}\mathbf{G}_{:,c}^{\mathrm{T}} & z^{-1}\mathbf{G}_{:,c} \\ z^{-1}\mathbf{G}_{:,c}^{\mathrm{T}} & z^{-1} \end{bmatrix} \\ &= \begin{bmatrix} \mathbf{Q}^{-1}[\ell] & \mathbf{0} \\ \mathbf{0} & 0 \end{bmatrix} + \frac{1}{z}\begin{bmatrix} \mathbf{G}_{:,c} \\ 1 \end{bmatrix}\begin{bmatrix} \mathbf{G}_{:,c}^{\mathrm{T}} & 1 \end{bmatrix}\end{aligned} \quad (22)$$

where

$$\begin{cases} \mathbf{G}_{:,c} = -\mathbf{Q}^{-1}[\ell]\boldsymbol{\eta}_{:,c} \\ z = Q_{c,c} - \boldsymbol{\eta}_{:,c}^{\mathrm{T}}\mathbf{Q}^{-1}[\ell]\boldsymbol{\eta}_{:,c} \end{cases}. \quad (23)$$

Therefore, computation of the inverse in the previous state can be reserved for the next state. The incremental forms of (18) and (19) respectively become

$$\begin{aligned}\mathbf{a}[\ell+1] &= (\mathbf{K}[\ell+1]+\rho\mathbf{I}[\ell+1])^{-1}(\mathbf{y}^{\mathrm{T}}[\ell+1]-b[\ell+1]\mathbf{e}^{\mathrm{T}}[\ell+1]) \\ &= \mathbf{Q}^{-1}[\ell+1](\mathbf{y}^{\mathrm{T}}[\ell+1]-b[\ell+1]\mathbf{e}^{\mathrm{T}}[\ell+1])\end{aligned} \quad (24)$$

and

$$b[\ell+1] = \frac{\mathbf{y}[\ell+1]\mathbf{Q}^{-1}[\ell+1]\mathbf{e}^{\mathrm{T}}[\ell+1]}{\mathbf{e}[\ell+1]\mathbf{Q}^{-1}[\ell+1]\mathbf{e}^{\mathrm{T}}[\ell+1]}. \quad (25)$$

For the decremental phase, given an index $r$ of a sample that is about to be removed, where $r \in \{1,\ldots,N\}$, we can rearrange the elements in $\mathbf{Q}^{-1}$, so that $r$ lies at the bottom-right corner of $\mathbf{Q}^{-1}$. Let $\boldsymbol{\Theta}$, $\boldsymbol{\xi}_r$, and $\theta_r$ respectively signify the three blocks of $\mathbf{Q}^{-1}$, shown in (26). Besides, $\boldsymbol{\Theta}$ is a matrix, $\boldsymbol{\xi}_r$ denotes a vector, and $\theta_r$ represents a scalar. Then,

$$\begin{aligned}\mathbf{Q}^{-1}[\ell] &= \begin{bmatrix} \boldsymbol{\Theta} & \boldsymbol{\xi}_r \\ \boldsymbol{\xi}_r^{\mathrm{T}} & \theta_r \end{bmatrix} \\ &= \begin{bmatrix} \mathbf{Q}^{-1}[\ell-1]+z^{-1}\mathbf{G}_{:,r}\mathbf{G}_{:,r}^{\mathrm{T}} & z^{-1}\mathbf{G}_{:,r} \\ z^{-1}\mathbf{G}_{:,r}^{\mathrm{T}} & z^{-1} \end{bmatrix}\end{aligned}. \quad (26)$$

The lower part of (26) comes from (22). Comparing the four blocks in the upper and lower parts of (22) [15] yields the following result.

$$\begin{aligned}\mathbf{Q}^{-1}[\ell-1] &= \boldsymbol{\Theta} - \frac{\boldsymbol{\xi}_r\boldsymbol{\xi}_r^{\mathrm{T}}}{\theta_r} \\ &= \mathbf{Q}^{-1}(1,1)[\ell] - \frac{\mathbf{Q}^{-1}(1,2)[\ell]\times\mathbf{Q}^{-1}(2,1)[\ell]}{\mathbf{Q}^{-1}(2,2)[\ell]}\end{aligned} \quad (27)$$

where $\mathbf{Q}^{-1}(\cdot,\cdot)$ indicates the blocks of $\mathbf{Q}^{-1}$. Substitution of (27) into (24) and (25) generates decremental forms.

### B. Multiple Incremental and Decremental Processes

Like Section II.B, also assume that the system adds $|C|$ new samples and removes $|R|$ existing data. For batch incremental learning, (22) becomes

$$\mathbf{Q}^{-1}[\ell+1] = \begin{bmatrix} \mathbf{Q}^{-1}[\ell]+\mathbf{G}_{:,C}\mathbf{Z}^{-1}\mathbf{G}_{:,C}^{\mathrm{T}} & \mathbf{G}_{:,C}\mathbf{Z}^{-\mathrm{T}} \\ \mathbf{Z}^{-1}\mathbf{G}_{:,C}^{\mathrm{T}} & \mathbf{Z}^{-1} \end{bmatrix} \quad (28)$$

where $\mathbf{Z}$ and $\mathbf{G}$ are matrices computed based on $|C|$ new samples. Notably, $\mathbf{Z}$ is a matrix version of $z$ in (23).

For batch decremental learning, (27) is replaced with (29).

$$\mathbf{Q}^{-1}[\ell-1] = \boldsymbol{\Theta} - \boldsymbol{\xi}_R\boldsymbol{\theta}_R^{-1}\boldsymbol{\xi}_R^{\mathrm{T}} \quad (29)$$

where $\boldsymbol{\xi}_R$ and $\boldsymbol{\theta}_R$ are computed based on $|R|$ decremental samples and the residual data. This step requires the inverse of $\boldsymbol{\theta}_R$. If the number of samples in $\mathbf{Q}^{-1}[\ell-1]$ is smaller than $|R|$, direct computation of $\mathbf{Q}^{-1}[\ell-1]$ saves more time.

To integrate multiple incremental/decremental processes together, the system should remove existing data first prior to adding new samples. Accordingly,

$$\mathbf{Q}^{-1}[\ell+1] = \begin{bmatrix} \mathbf{\Theta} - \xi_R \theta_R^{-1} \xi_R^T + \mathbf{G}_{:,C} \mathbf{Z}^{-1} \mathbf{G}_{:,C}^T & \mathbf{G}_{:,C} \mathbf{Z}^{-T} \\ \mathbf{Z}^{-1} \mathbf{G}_{:,C}^T & \mathbf{Z}^{-1} \end{bmatrix}. \quad (30)$$

## IV. Incremental/Decremental Kernelized Bayesian Regression

Unlike KRR that focuses on frequentist methodologies, where sufficient occurrences are observed, Bayesian Regression concentrates on uncertainty modeling. Thus, statistical distributions are introduced in Kernelized Bayesian Regression (KBR). The observed instances are sampled from a stochastic process that fits a statistical distribution. As Bayesian theory works for various distributions, this work uses Gaussian distributions as a case study for modeling incremental and decremental analysis.

To avoid confusion, this study uses the following notations to describe the relation between independent variables (i.e., predictors) and their conditional parameters in the subsequent functions.

- $P(\cdot|\cdot)$: When the transposition operator appears in the independent variable (i.e., the first operant) of a probabilistic function, the conditional parameters (i.e., the subsequent operants) still remain their original notations without adding the transposition operator. Notably, $P(\cdot^T|\cdot) \neq P(\cdot|\cdot)$.
- $\mathcal{N}(\cdot|\cdot)$: When the transpose operator is used in the independent variable of a normal distribution function, the subsequent hyperparameters reflect such a change and use the transpose operator.
- $\Sigma_{\cdot|\cdot}$ or $\mathbf{\Sigma}_{\cdot|\cdot}$: The first operant in the subscript is viewed as the independent argument of a covariance matrix function. When there is a transposition operation, the operant displays such an operator in the notation, e.g., $\mathbf{\Sigma}_{\mathbf{y}^T|\mathbf{u},\mathbf{\Phi}}$. The conditional operant in the subscript remains the same form unless it is a dependent response of regression, e.g., $\mathbf{\Sigma}_{\mathbf{u}|\mathbf{y}^T,\mathbf{\Phi}}$.
- $\mu_{\cdot|\cdot}$ or $\boldsymbol{\mu}_{\cdot|\cdot}$: They are based on the above-mentioned notations.

Moreover, for uncertainty modeling, $b$ in (1) should be changed to a random variable $b_i$ corresponding to its observed sample $(\mathbf{x}_i, y_i)$. Consider a regression model,

$$y_i = \mathbf{u}^T \phi(\mathbf{x}_i) + b_i, \quad (31)$$

or in a matrix form,

$$\mathbf{y} = \mathbf{u}^T \mathbf{\Phi} + \mathbf{b}$$

where $P(b) \sim \mathcal{N}(\mu_\mathbf{b}, \Sigma_\mathbf{b})$ and $P(\phi(\mathbf{x}_i) \in \mathbf{\Phi}) \sim \mathcal{N}(\boldsymbol{\mu}_\mathbf{\Phi}, \mathbf{\Sigma}_\mathbf{\Phi})$. Besides, $\mu_\mathbf{b}$ and $\Sigma_\mathbf{b}$ are scalars, and the dimensions of $\boldsymbol{\mu}_\mathbf{\Phi}$ and $\mathbf{\Sigma}_\mathbf{\Phi}$ are $J$-by-1 and $J$-by-$J$, respectively. Furthermore, for simplicity, assume $\mu_\mathbf{b} = 0$. Also, assume $\mathbf{x}_i$ and $b_i$ are independent. Thus,

$$\mu_\mathbf{y} = \mathbf{u}^T \boldsymbol{\mu}_\mathbf{\Phi} + \mu_\mathbf{b} \\ = \mathbf{u}^T \boldsymbol{\mu}_\mathbf{\Phi} \quad (32)$$

and

$$\mathbf{\Sigma}_\mathbf{y} = \mathbf{u}^T \mathbf{\Sigma}_\mathbf{\Phi} \mathbf{u} + \Sigma_\mathbf{b} \\ = \mathbf{u}^T \mathbf{\Sigma}_\mathbf{\Phi} \mathbf{u} + \sigma_\mathbf{b}^2. \quad (33)$$

Furthermore, $\mu_\mathbf{y}$ and $\Sigma_\mathbf{y}$ are scalars, and $\mathbf{\Sigma}_{\mathbf{\Phi}\mathbf{y}} = \mathbf{\Sigma}_\mathbf{\Phi} \times \mathbf{u}$. The sample mean and the sample covariance matrix, $\boldsymbol{\mu}_\mathbf{\Phi}$ and $\mathbf{\Sigma}_\mathbf{\Phi}$, are respectively

$$\boldsymbol{\mu}_\mathbf{\Phi} = \frac{1}{N} \sum_{i=1}^{N} \phi(\mathbf{x}_i)$$

and

$$\mathbf{\Sigma}_\mathbf{\Phi} = \frac{1}{N-1} \sum_{i=1}^{N} (\phi(\mathbf{x}_i) - \boldsymbol{\mu}_\mathbf{\Phi})(\phi(\mathbf{x}_i) - \boldsymbol{\mu}_\mathbf{\Phi})^T.$$

For homoscedasticity, this study assumes that the covariance between residues (i.e., $b_i$) are the same. The intrinsic covariance matrix and the empirical covariance matrix are respectively

$$\Sigma_\mathbf{b} = E\left[(\mathbf{b} - \boldsymbol{\mu}_\mathbf{b})(\mathbf{b} - \boldsymbol{\mu}_\mathbf{b})^T\right] = \sigma_\mathbf{b}^2$$

and

$$\mathbf{\Psi}_\mathbf{b} = \begin{bmatrix} \sigma_1^2 & 0 & 0 \\ 0 & \ddots & 0 \\ 0 & 0 & \sigma_N^2 \end{bmatrix} = \begin{bmatrix} \sigma_\mathbf{b}^2 & 0 & 0 \\ 0 & \ddots & 0 \\ 0 & 0 & \sigma_\mathbf{b}^2 \end{bmatrix} = \sigma_\mathbf{b}^2 \mathbf{I}.$$

Consequently,

$$\mathbf{\Sigma}_{\mathbf{b}^T} = \mathbf{\Psi}_\mathbf{b} = \Sigma_\mathbf{b} \mathbf{I}.$$

This equation is subsequently used in (40).

### A. Training Stage

In Bayesian inference, prior information serves as a model or function parameters to interpret the likelihood probability of observed data $(\mathbf{x}_i, y_i)$. The training stage uses Bayesian inference to estimate the posterior distribution of $\mathbf{u}$, which consists of two parts. One is the likelihood probability "$P(\mathbf{y}^T|\mathbf{u},\mathbf{\Phi})$," and the other is the prior probability "$P(\mathbf{u})$."

It is worth noting that the dimensions of two independent variables should fit their joint probability, i.e., $P([\mathbf{u} \ \mathbf{y}^T])$. Therefore, the transpose of $\mathbf{y}$ is used herein.

$$P(\mathbf{u}|\boldsymbol{\Phi},\mathbf{y}) = \frac{P(\mathbf{y}^T|\mathbf{u},\boldsymbol{\Phi})P(\mathbf{u})}{P(\mathbf{y}^T|\boldsymbol{\Phi})} \quad (34)$$
$$\propto P(\mathbf{y}^T|\mathbf{u},\boldsymbol{\Phi})P(\mathbf{u})$$

where the marginal likelihood is

$$P(\mathbf{y}^T|\boldsymbol{\Phi}) = \int P(\mathbf{y}^T|\mathbf{u},\boldsymbol{\Phi})P(\mathbf{u})d\mathbf{u}. \quad (35)$$

The following steps establish the posterior distribution by computing the likelihood and prior probabilities.

■ Computation of the Likelihood Probability

As $P(b) \sim \mathcal{N}(0,\sigma_b^2)$ and $P(\phi(\mathbf{x}_i)) \sim \mathcal{N}(\boldsymbol{\mu}_\Phi,\boldsymbol{\Sigma}_\Phi)$, the Gaussian distribution of the likelihood $P(\mathbf{y}|\mathbf{u},\boldsymbol{\Phi})$ is $\mathcal{N}(\boldsymbol{\mu}_{\mathbf{y}|\mathbf{u},\boldsymbol{\Phi}},\boldsymbol{\Sigma}_{\mathbf{y}|\mathbf{u},\boldsymbol{\Phi}}) \sim \mathcal{N}(\mathbf{u}^T\boldsymbol{\Phi},\sigma_b^2)$ based on the following conditional expectation and conditional covariance of a linear Gaussian system [26]. That is,

$$\boldsymbol{\Sigma}_{\mathbf{y}|\boldsymbol{\Phi}} = \boldsymbol{\Sigma}_\mathbf{y} - \boldsymbol{\Sigma}_{\mathbf{y}\boldsymbol{\Phi}}\boldsymbol{\Sigma}_\Phi^{-1}\boldsymbol{\Sigma}_{\boldsymbol{\Phi}\mathbf{y}} \quad (36)$$
$$= \sigma_b^2$$

and

$$\boldsymbol{\mu}_{\mathbf{y}|\boldsymbol{\Phi}} = \boldsymbol{\mu}_\mathbf{y} + \boldsymbol{\Sigma}_{\mathbf{y}\boldsymbol{\Phi}}\boldsymbol{\Sigma}_\Phi^{-1}(\boldsymbol{\Phi} - \boldsymbol{\mu}_\Phi \mathbf{e})$$
$$= \mathbf{u}^T\boldsymbol{\Phi} + \mu_b \quad (37)$$
$$= \mathbf{u}^T\boldsymbol{\Phi}$$

where $\boldsymbol{\mu}_{\mathbf{y}|\boldsymbol{\Phi}} = \mu_{y|\Phi} \times \mathbf{e}$, $\boldsymbol{\mu}_\mathbf{y} = \mu_y \times \mathbf{e}$, and $\boldsymbol{\mu}_\mathbf{b} = \mu_b \times \mathbf{e}$. Besides, $\mu_{y|\Phi}$ and $\Sigma_{y|\Phi}$ are scalars, and $\boldsymbol{\mu}_{\mathbf{y}|\boldsymbol{\Phi}}$ is a 1-by-$N$ vector. Let $\mathrm{tr}(\cdot)$ denote trace operations and $\det(\cdot)$ represent the determinant. Accordingly,

$$P(\mathbf{y}|\mathbf{u},\boldsymbol{\Phi})$$
$$= \prod_{i=1}^{N} P(y_i|\mathbf{u},\boldsymbol{\Phi})$$
$$= \frac{1}{\sqrt{2\pi\cdot\det(\Sigma_{\mathbf{y}|\mathbf{u},\boldsymbol{\Phi}})}}\exp\left(-\frac{1}{2}\sum_i(y_i - \boldsymbol{\mu}_{\mathbf{y}|\mathbf{u},\boldsymbol{\Phi}})^T \Sigma_{\mathbf{y}|\mathbf{u},\boldsymbol{\Phi}}^{-1}(y_i - \boldsymbol{\mu}_{\mathbf{y}|\mathbf{u},\boldsymbol{\Phi}})\right)$$
$$= \frac{1}{\left(\sqrt{2\pi\cdot\det(\Sigma_{\mathbf{y}|\mathbf{u},\boldsymbol{\Phi}})}\right)^N}\exp\left(-\frac{1}{2}\mathrm{tr}\left((\mathbf{y} - \boldsymbol{\mu}_{\mathbf{y}|\mathbf{u},\boldsymbol{\Phi}})^T \Sigma_{\mathbf{y}|\mathbf{u},\boldsymbol{\Phi}}^{-1}(\mathbf{y} - \boldsymbol{\mu}_{\mathbf{y}|\mathbf{u},\boldsymbol{\Phi}})\right)\right)$$
$$= \frac{1}{\left(\sqrt{2\pi\cdot\det(\sigma_b^2)}\right)^N}\exp\left(-\frac{1}{2}\mathrm{tr}\left((\mathbf{y} - \mathbf{u}^T\boldsymbol{\Phi})^T(\sigma_b^{-2})(\mathbf{y} - \mathbf{u}^T\boldsymbol{\Phi})\right)\right)$$
$$= \frac{1}{\left(\sqrt{2\pi\sigma_b^2}\right)^N}\exp\left(-\frac{1}{2\sigma_b^2}\mathrm{tr}\left((\mathbf{y} - \mathbf{u}^T\boldsymbol{\Phi})^T(\mathbf{y} - \mathbf{u}^T\boldsymbol{\Phi})\right)\right).$$
$$(38)$$

■ Computation of the Prior Probability

Theoretically, the prior probability can be any distribution. However, such selection would result in posterior distributions without analytical solutions [27]. This benefits no computation. As the likelihood probability is a Gaussian distribution, we can use conjugate prior distributions to model the system for convenience of computation. When the generated posterior distribution and the selected prior distribution belong to the same class of distributions, such a prior distribution is a conjugate prior distribution [28]. There is a systematic analytical model for conjugate prior distributions [29].

To generate a Gaussian posterior distribution, this study selects a Gaussian prior distribution (39) for $P(\mathbf{u}) \sim \mathcal{N}(\boldsymbol{\mu}_\mathbf{u},\boldsymbol{\Sigma}_\mathbf{u})$. The parameters, $\boldsymbol{\mu}_\mathbf{u}$ and $\boldsymbol{\Sigma}_\mathbf{u}$, can be set to $\mathbf{0}$ and $\sigma_\mathbf{u}^2\mathbf{I}$, respectively, for simplicity. The dimensions of them are, respectively, $J$-by-1 and $J$-by-$J$.

$$P(\mathbf{u}) = \frac{1}{\sqrt{(2\pi)^J \cdot \det(\boldsymbol{\Sigma}_\mathbf{u})}}\exp\left(-\frac{1}{2}\mathrm{tr}\left[(\mathbf{u}-\boldsymbol{\mu}_\mathbf{u})^T \boldsymbol{\Sigma}_\mathbf{u}^{-1}(\mathbf{u}-\boldsymbol{\mu}_\mathbf{u})\right]\right).$$
$$(39)$$

■ Computation of the Posterior Probability

It is worth noting that the likelihood $P(\mathbf{y}|\mathbf{u},\boldsymbol{\Phi})$ is $\mathcal{N}(\boldsymbol{\mu}_{\mathbf{y}|\mathbf{u},\boldsymbol{\Phi}} = \mathbf{u}^T\boldsymbol{\Phi}, \Sigma_{\mathbf{y}|\mathbf{u},\boldsymbol{\Phi}} = \sigma_b^2)$, and the prior probability $P(\mathbf{u})$ is $\mathcal{N}(\boldsymbol{\mu}_\mathbf{u},\boldsymbol{\Sigma}_\mathbf{u})$. According to Gaussian identities in [20, 28-30], the posterior distribution is Gaussian.

To fit the joint probability of $\mathbf{y}$ and $\mathbf{u}$, plugging $P(\mathbf{y}^T|\mathbf{u},\boldsymbol{\Phi}) \sim \mathcal{N}(\mathbf{y}^T|\boldsymbol{\mu}_{\mathbf{y}^T|\mathbf{u},\boldsymbol{\Phi}},\boldsymbol{\Sigma}_{\mathbf{y}^T|\mathbf{u},\boldsymbol{\Phi}}) = \mathcal{N}(\mathbf{y}^T|\boldsymbol{\mu}_{\mathbf{y}^T|\mathbf{u},\boldsymbol{\Phi}},\boldsymbol{\Psi}_{\mathbf{y}|\mathbf{u},\boldsymbol{\Phi}})$ into (34) yields

$$P(\mathbf{u}|\boldsymbol{\Phi},\mathbf{y}) \propto P(\mathbf{y}^T|\mathbf{u},\boldsymbol{\Phi})P(\mathbf{u})$$
$$= \mathcal{N}(\mathbf{y}^T|\boldsymbol{\mu}_{\mathbf{y}^T|\mathbf{u},\boldsymbol{\Phi}},\boldsymbol{\Sigma}_{\mathbf{y}^T|\mathbf{u},\boldsymbol{\Phi}})\mathcal{N}(\mathbf{u}|\boldsymbol{\mu}_\mathbf{u},\boldsymbol{\Sigma}_\mathbf{u})$$
$$= \mathcal{N}(\mathbf{y}^T|\boldsymbol{\Phi}^T\mathbf{u},\sigma_b^2\mathbf{I})\mathcal{N}(\mathbf{u}|\boldsymbol{\mu}_\mathbf{u},\boldsymbol{\Sigma}_\mathbf{u}) \quad (40)$$
$$\propto \mathcal{N}(\mathbf{u}|\boldsymbol{\mu}_{\mathbf{u}|\mathbf{y}^T,\boldsymbol{\Phi}},\boldsymbol{\Sigma}_{\mathbf{u}|\mathbf{y}^T,\boldsymbol{\Phi}})$$

where

$$\boldsymbol{\Sigma}_{\mathbf{u}|\mathbf{y}^T,\boldsymbol{\Phi}} = \boldsymbol{\Sigma}_\mathbf{u} - \boldsymbol{\Sigma}_\mathbf{u}\boldsymbol{\Phi}(\boldsymbol{\Phi}^T\boldsymbol{\Sigma}_\mathbf{u}\boldsymbol{\Phi} + \boldsymbol{\Sigma}_{\mathbf{b}^T})^{-1}\boldsymbol{\Phi}^T\boldsymbol{\Sigma}_\mathbf{u} \quad (41)$$
$$= (\boldsymbol{\Sigma}_\mathbf{u}^{-1} + \sigma_b^{-2}\boldsymbol{\Phi}\boldsymbol{\Phi}^T)^{-1}$$

and

$$\begin{aligned}
\boldsymbol{\mu}_{\mathbf{u}|\mathbf{y}^\mathrm{T},\boldsymbol{\Phi}} &= \boldsymbol{\mu}_\mathbf{u} + \boldsymbol{\Sigma}_\mathbf{u}\boldsymbol{\Phi}\left(\boldsymbol{\Phi}^\mathrm{T}\boldsymbol{\Sigma}_\mathbf{u}\boldsymbol{\Phi}+\boldsymbol{\Sigma}_{\mathbf{b}^\mathrm{T}}\right)^{-1}\left(\mathbf{y}^\mathrm{T}-\boldsymbol{\mu}_\mathbf{y}^\mathrm{T}\right)\\
&= \boldsymbol{\mu}_\mathbf{u} + \left(\boldsymbol{\Sigma}_\mathbf{u}^{-1}+\boldsymbol{\Phi}\boldsymbol{\Sigma}_{\mathbf{b}^\mathrm{T}}^{-1}\boldsymbol{\Phi}^\mathrm{T}\right)^{-1}\boldsymbol{\Phi}\boldsymbol{\Sigma}_{\mathbf{b}^\mathrm{T}}^{-1}\left(\mathbf{y}^\mathrm{T}-\boldsymbol{\mu}_\mathbf{y}^\mathrm{T}\right)\\
&= \boldsymbol{\mu}_\mathbf{u} + \boldsymbol{\Sigma}_{\mathbf{u}|\mathbf{y}^\mathrm{T}}\boldsymbol{\Phi}\boldsymbol{\Sigma}_{\mathbf{b}^\mathrm{T}}^{-1}\left(\mathbf{y}^\mathrm{T}-\boldsymbol{\mu}_\mathbf{y}^\mathrm{T}\right)\\
&= \boldsymbol{\Sigma}_{\mathbf{u}|\mathbf{y}^\mathrm{T}}\left(\boldsymbol{\Sigma}_{\mathbf{u}|\mathbf{y}^\mathrm{T}}^{-1}\boldsymbol{\mu}_\mathbf{u}+\boldsymbol{\Phi}\boldsymbol{\Sigma}_{\mathbf{b}^\mathrm{T}}^{-1}\left(\mathbf{y}^\mathrm{T}-\boldsymbol{\mu}_\mathbf{y}^\mathrm{T}\right)\right)\\
&= \boldsymbol{\Sigma}_{\mathbf{u}|\mathbf{y}^\mathrm{T},\boldsymbol{\Phi}}\left(\boldsymbol{\Sigma}_\mathbf{u}^{-1}\boldsymbol{\mu}_\mathbf{u}+\sigma_\mathbf{b}^{-2}\boldsymbol{\Phi}\left(\mathbf{y}-\mathbf{b}\right)^\mathrm{T}\right)
\end{aligned} \quad (42)$$

Moreover, $\boldsymbol{\mu}_{\mathbf{u}|\mathbf{y}^\mathrm{T},\boldsymbol{\Phi}}$ is a $J$-by-1 vector, and $\boldsymbol{\Sigma}_{\mathbf{u}|\mathbf{y}^\mathrm{T},\boldsymbol{\Phi}}$ is a $J$-by-$J$ matrix. Assume that the prior information changes with time. Subsequently,

$$\begin{aligned}
&\boldsymbol{\Sigma}_{\mathbf{u}|\mathbf{y}^\mathrm{T},\boldsymbol{\Phi}}[\ell+1]\\
&= \left(\boldsymbol{\Sigma}_{\mathbf{u}|\mathbf{y}^\mathrm{T},\boldsymbol{\Phi}}^{-1}[\ell]+\sigma_\mathbf{b}^{-2}\left(\boldsymbol{\Phi}\boldsymbol{\Phi}^\mathrm{T}\right)[\ell]\right)^{-1}\\
&= \boldsymbol{\Sigma}_{\mathbf{u}|\mathbf{y}^\mathrm{T},\boldsymbol{\Phi}}[\ell]-\sigma_\mathbf{b}^{-2}\boldsymbol{\Sigma}_{\mathbf{u}|\mathbf{y}^\mathrm{T},\boldsymbol{\Phi}}[\ell]\times\\
&\quad \left(\mathbf{I}+\sigma_\mathbf{b}^{-2}\left(\boldsymbol{\Phi}\boldsymbol{\Phi}^\mathrm{T}\right)[\ell]\boldsymbol{\Sigma}_{\mathbf{u}|\mathbf{y}^\mathrm{T},\boldsymbol{\Phi}}[\ell]\right)^{-1}\left(\boldsymbol{\Phi}\boldsymbol{\Phi}^\mathrm{T}\right)[\ell]\boldsymbol{\Sigma}_{\mathbf{u}|\mathbf{y}^\mathrm{T},\boldsymbol{\Phi}}[\ell]
\end{aligned} \quad (43)$$

where

$$\left(\boldsymbol{\Phi}\boldsymbol{\Phi}^\mathrm{T}\right)[\ell]=\left(\boldsymbol{\Phi}\boldsymbol{\Phi}^\mathrm{T}\right)[\ell-1]+\boldsymbol{\Phi}_\mathcal{H}\boldsymbol{\Phi}'_\mathcal{H}$$

and

$$\begin{aligned}
&\boldsymbol{\mu}_{\mathbf{u}|\mathbf{y}^\mathrm{T},\boldsymbol{\Phi}}[\ell+1]\\
&=\boldsymbol{\Sigma}_{\mathbf{u}|\mathbf{y}^\mathrm{T},\boldsymbol{\Phi}}[\ell+1]\left(\boldsymbol{\Sigma}_{\mathbf{u}|\mathbf{y}^\mathrm{T},\boldsymbol{\Phi}}^{-1}[\ell]\boldsymbol{\mu}_{\mathbf{u}|\mathbf{y}^\mathrm{T},\boldsymbol{\Phi}}[\ell]+\right.\\
&\quad \left.\sigma_\mathbf{b}^{-2}\boldsymbol{\Phi}[\ell]\left(\mathbf{y}^\mathrm{T}[\ell]-\mathbf{b}^\mathrm{T}[\ell]\right)\right).
\end{aligned} \quad (44)$$

When existing training samples change, $\boldsymbol{\Phi}\boldsymbol{\Phi}^\mathrm{T}$ and $\boldsymbol{\Phi}\mathbf{y}^\mathrm{T}$ in (43) and (44), respectively, should be accordingly updated. Otherwise, only prior information is updated. The posterior probability reflects the modification in training samples and prior information.

### B. Predictive Stage

As the training stage already generates the posterior distribution and the uncertainty of $\mathbf{u}$, the posterior predictive distribution "$P((y^*)^\mathrm{T}|\phi(\mathbf{x}^*),\boldsymbol{\Phi},\mathbf{y})$" is then used to model the uncertainty of the predictive output. The posterior predictive distribution can be rewritten as the marginal distribution of "$P((y^*)^\mathrm{T}|\phi(\mathbf{x}^*),\mathbf{u})$" and "$P(\mathbf{u}|\boldsymbol{\Phi},\mathbf{y})$." Let $y^*$ represent the scalar predictive output of the model when an $M$-by-1 testing sample $\mathbf{x}^*$ is input. Applying the product rule and the integral rule of Gaussian identities [30] to the marginal distribution yields the following form.

$$\begin{aligned}
&P\left((y^*)^\mathrm{T}|\phi(\mathbf{x}^*),\boldsymbol{\Phi},\mathbf{y}\right)\\
&=\int P\left((y^*)^\mathrm{T}|\phi(\mathbf{x}^*),\mathbf{u}\right)P(\mathbf{u}|\boldsymbol{\Phi},\mathbf{y})d\mathbf{u}\\
&=\int \mathcal{N}\left((y^*)^\mathrm{T}|\phi(\mathbf{x}^*)^\mathrm{T}\mathbf{u},\sigma_\mathbf{b}^2\right)\mathcal{N}\left(\mathbf{u}|\boldsymbol{\mu}_{\mathbf{u}|\mathbf{y},\boldsymbol{\Phi}},\boldsymbol{\Sigma}_{\mathbf{u}|\mathbf{y},\boldsymbol{\Phi}}\right)d\mathbf{u}\\
&\propto \mathcal{N}\left((y^*)^\mathrm{T}|\phi(\mathbf{x}^*)^\mathrm{T}\boldsymbol{\mu}_{\mathbf{u}|\mathbf{y},\boldsymbol{\Phi}},\sigma_\mathbf{b}^2+\phi(\mathbf{x}^*)^\mathrm{T}\boldsymbol{\Sigma}_{\mathbf{u}|\mathbf{y},\boldsymbol{\Phi}}\phi(\mathbf{x}^*)\right)\\
&=\mathcal{N}\left((y^*)^\mathrm{T}|\mu^*,\Psi^*\right).
\end{aligned} \quad (45)$$

Notably, although $y^*$ is a scalar, (45) still uses the transpose of $y^*$ for clarity. If $P((y^*)^\mathrm{T}|\phi(\mathbf{x}^*),\mathbf{u})$ follows the distribution of the training data $P(\mathbf{y}^\mathrm{T}|\mathbf{u},\boldsymbol{\Phi})$ due to the need for analytical solutions, the predictive distribution becomes Gaussian. Accordingly,

$$\begin{aligned}
&P\left((y^*)^\mathrm{T}|\phi(\mathbf{x}^*),\boldsymbol{\Phi},\mathbf{y}\right)\\
&=\frac{1}{\sqrt{2\pi\cdot\det(\Psi^*)}}\exp\left(-\frac{1}{2}\mathrm{tr}\left(\left((y^*)^\mathrm{T}-\mu^*\right)^\mathrm{T}(\Psi^*)^{-1}\left((y^*)^\mathrm{T}-\mu^*\right)\right)\right)
\end{aligned} \quad (46)$$

where

$$\Psi^*=\sigma_\mathbf{b}^2+\phi(\mathbf{x}^*)^\mathrm{T}\boldsymbol{\Sigma}_{\mathbf{u}|\mathbf{y},\boldsymbol{\Phi}}\phi(\mathbf{x}^*) \quad (47)$$

and

$$\mu^*=\phi(\mathbf{x}^*)^\mathrm{T}\boldsymbol{\mu}_{\mathbf{u}|\mathbf{y},\boldsymbol{\Phi}}. \quad (48)$$

Moreover, $\mu^*$ and $\Psi^*$ are scalars. When existing training samples change, $\boldsymbol{\Sigma}_{\mathbf{u}|\mathbf{y},\boldsymbol{\Phi}}$ in (47) and $\boldsymbol{\mu}_{\mathbf{u}|\mathbf{y},\boldsymbol{\Phi}}$ in (48) need updates, respectively. Subsequently, a new posterior predictive distribution is generated.

When variable prior information is involved in the update, (47) and (48) become

$$\Psi^*=\sigma_\mathbf{b}^2+\phi(\mathbf{x}^*)^\mathrm{T}\boldsymbol{\Sigma}_{\mathbf{u}|\mathbf{y},\boldsymbol{\Phi}}[\ell]\phi(\mathbf{x}^*) \quad (49)$$

and

$$\mu^*=\phi(\mathbf{x}^*)^\mathrm{T}\boldsymbol{\mu}_{\mathbf{u}|\mathbf{y},\boldsymbol{\Phi}}[\ell]. \quad (50)$$

## V. Experimental Results

Experiments on open datasets were carried out for evaluating the performance. The information of these datasets is listed in Table I. The first column shows the name. The rest columns specify the number of classes, samples, and dimensions, respectively. Dataset "MIT/BIH ECG" is available at PhysioNet (www.physionet.org), and "Dorothea (DRT)" was downloaded from the UC Irvine (UCI) Machine Learning Repository (archive.ics.uci.edu/ml/). The datasets show two typical data, where both $N > M$ and $M > N$ are presented, respectively.

The experiment used approximately 80.00% of the data for training and 20.00% of the data for testing. Furthermore, +4 and −2 samples were randomly selected for incremental and decremental computation at the same time. Table II and Table III summarize the incremental and decremental settings. The algorithmic details are listed in Table III. Ridges were empirically set in the experiments. Regarding KBR settings, $\mu_u$ and $\mu_b$ were **0** and 0, respectively. Besides, both $\sigma_u^2$ and $\sigma_b^2$ were set to 0.01.

Table I
Attributes of the Datasets

| Name | #Classes | #Samples | #Dimensions |
|---|---|---|---|
| ECG | 2 | 104033 | 21 |
| DRT | 2 | 800 | 1000000 |

Table II
Settings of Incremental/Decremental Computation

| Name | Basic Training Size | Multiple Incremental/Decremental Size |
|---|---|---|
| ECG | 83226 | +4 / −2 |
| DRT | 640 | +4 / −2 |

Table III
Algorithmic Settings

| Name | Kernel | Ridge |
|---|---|---|
| Intrinsic-Space KRR | Poly2 & Poly3 | 0.5 |
| Empirical-Space KRR | Poly2, Poly3, & RBF | 0.5 |

*RBFs are inapplicable to intrinsic space due to infinite dimensions. The radius of RBFs is 50.00.

Two baselines, "nonincremental analysis" and "single incremental algorithm," along with the proposed method "multiple incremental approach" were used for comparison. In total, ten rounds of data operations (i.e., data insertion and deletion) were evaluated, and computational time in log10 was calculated. For the nonincremental part, it recomputed the weight of the system based on the new dataset after one round of data operations. Regarding the single incremental part, it reanalyzed the new dataset every time when data insertion or deletion occurred. Besides, only when one round of data operations was complete, cumulative computational time was measured.

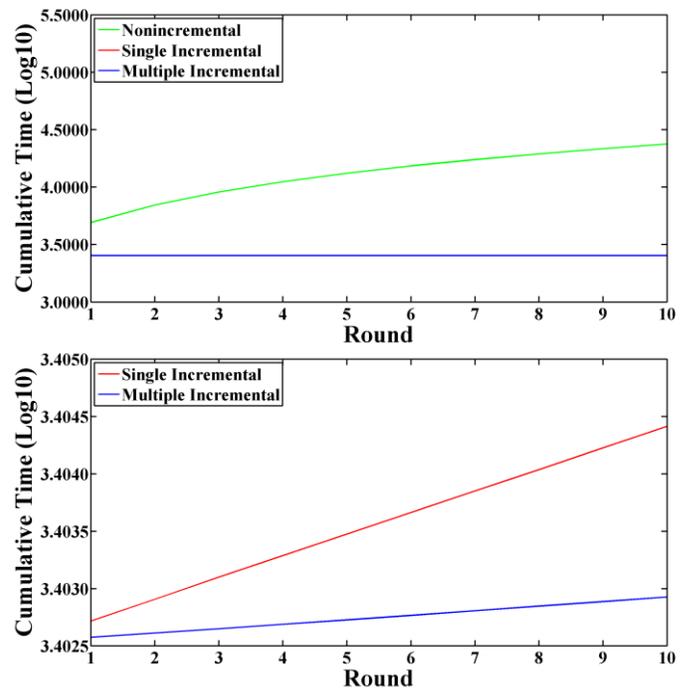

Fig. 2. KRR comparison between multiple incremental (blue), single incremental (red) and nonincremental (green) learning with the use of the ECG dataset and the poly2 kernel. The computational time (log10) was cumulative. The accuracy rates were all 94.71%.

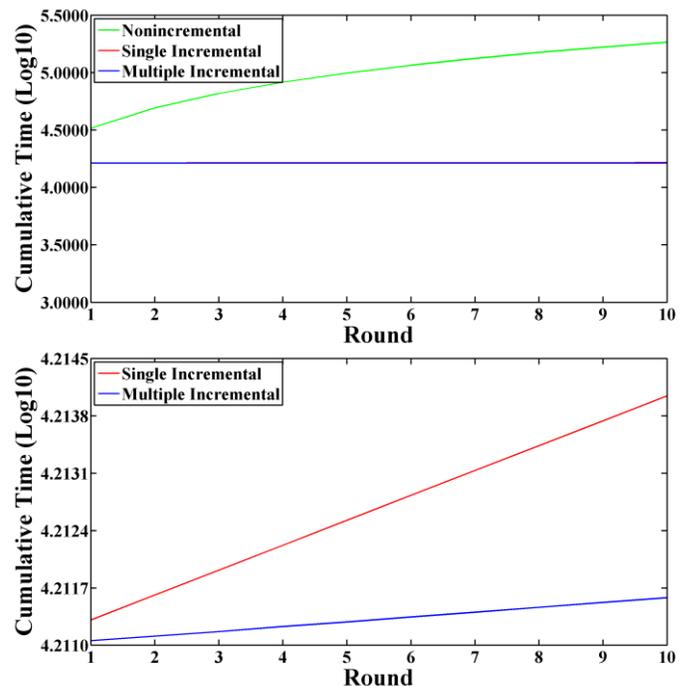

Fig. 3. KRR comparison between multiple incremental (blue), single incremental (red) and nonincremental (green) learning with the use of the ECG dataset and the poly3 kernel. The computational time (log10) was cumulative. The accuracy rates were all 97.37%.

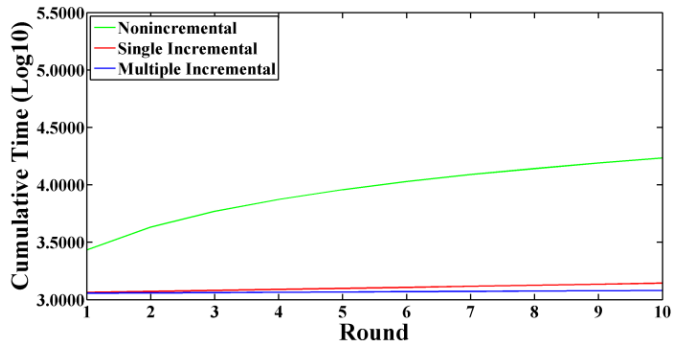

Fig. 4. KRR comparison between multiple incremental (blue), single incremental (red) and nonincremental (green) learning with the use of the DRT dataset and the poly2 kernel. The computational time (log10) was cumulative. The accuracy rates were all 90.00%.

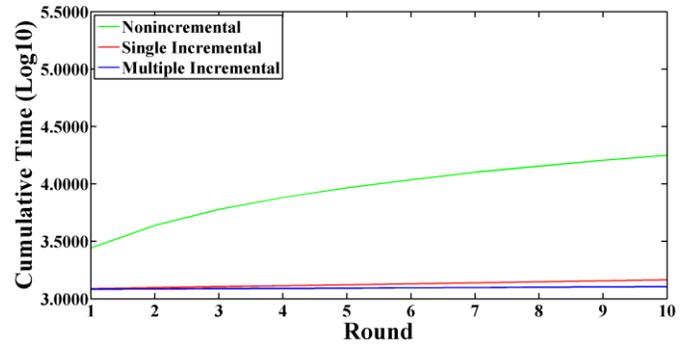

Fig. 5. KRR comparison between multiple incremental (blue), single incremental (red) and nonincremental (green) learning with the use of the DRT dataset and the poly3 kernel. The computational time (log10) was cumulative. The accuracy rates were all 90.00%.

Table IV
KRR Computational Time (log10) Based on the ECG Dataset and the Poly2 Kernel in a Single Round

| #Samples | 83226 | 83228 | 83230 | 83232 | 83234 | 83236 | 83238 | 83240 | 83242 | 83244 |
|---|---|---|---|---|---|---|---|---|---|---|
| Multiple | -0.537544 | -0.665259 | -0.659984 | -0.635436 | -0.651824 | -0.645394 | -0.634669 | -0.622588 | -0.643913 | -0.623469 |
| Single | 0.047783 | 0.043765 | 0.050801 | 0.038683 | 0.040046 | 0.041661 | 0.039198 | 0.036630 | 0.042320 | 0.041475 |
| None | 3.376356 | 3.314288 | 3.316463 | 3.317598 | 3.315914 | 3.317286 | 3.317168 | 3.317430 | 3.326118 | 3.331818 |

Table V
KRR Computational Time (log10) Based on the ECG Dataset and the Poly3 Kernel in a Single Round

| #Samples | 83226 | 83228 | 83230 | 83232 | 83234 | 83236 | 83238 | 83240 | 83242 | 83244 |
|---|---|---|---|---|---|---|---|---|---|---|
| Multiple | 4.211003 | 0.297297 | 0.314129 | 0.314672 | 0.364845 | 0.361343 | 0.334149 | 0.354265 | 0.340383 | 0.337915 |
| Single | 4.224946 | 1.058435 | 1.056797 | 1.056978 | 1.056486 | 1.059564 | 1.058412 | 1.055703 | 1.057832 | 1.061875 |
| None | 4.211003 | 4.214649 | 4.214517 | 4.219702 | 4.219394 | 4.224266 | 4.226119 | 4.230048 | 4.226973 | 4.241862 |

Table VI
KRR Computational Time (log10) Based on the DRT Dataset and the Poly2 Kernel in a Single Round

| #Samples | 640 | 642 | 644 | 646 | 648 | 650 | 652 | 654 | 656 | 658 |
|---|---|---|---|---|---|---|---|---|---|---|
| Multiple | 3.053674 | 0.846649 | 0.720064 | 0.850986 | 0.845865 | 0.853454 | 0.851205 | 0.855350 | 0.856517 | 0.797779 |
| Single | 3.051355 | 1.373776 | 1.351769 | 1.373161 | 1.400000 | 1.426793 | 1.422169 | 1.445650 | 1.453737 | 1.452745 |
| None | 3.053674 | 3.196123 | 3.196231 | 3.201359 | 3.201729 | 3.206425 | 3.211160 | 3.178982 | 3.217578 | 3.217389 |

Table VII
KRR Computational Time (log10) Based on the DRT Dataset and the Poly3 Kernel in a Single Round

| #Samples | 640 | 642 | 644 | 646 | 648 | 650 | 652 | 654 | 656 | 658 |
|---|---|---|---|---|---|---|---|---|---|---|
| Multiple | 0.853478 | 0.718077 | 0.856490 | 0.878286 | 0.862960 | 0.851657 | 0.903701 | 0.898904 | 0.901134 | 0.841226 |
| Single | 1.373330 | 1.348596 | 1.371429 | 1.393420 | 1.406572 | 1.424955 | 1.421473 | 1.444416 | 1.459492 | 1.454026 |
| None | 3.194155 | 3.198641 | 3.214183 | 3.208231 | 3.209122 | 3.213412 | 3.247027 | 3.212538 | 3.250765 | 3.228786 |

Table VIII
KRR Computational Time (log10) Based on the DRT Dataset and the RBF in a Single Round

| #Samples | 640 | 642 | 644 | 646 | 648 | 650 | 652 | 654 | 656 | 658 |
|---|---|---|---|---|---|---|---|---|---|---|
| Multiple | 0.888406 | 0.776181 | 0.852696 | 0.851705 | 0.848764 | 0.853636 | 0.852904 | 0.854650 | 0.858440 | 0.801611 |
| Single | 1.419054 | 1.394077 | 1.419183 | 1.439993 | 1.450303 | 1.468781 | 1.466268 | 1.478293 | 1.485936 | 1.487907 |
| None | 3.225958 | 3.218848 | 3.201681 | 3.206244 | 3.208604 | 3.207531 | 3.210940 | 3.179368 | 3.217160 | 3.218175 |

Table IX
KRR Average Computational Time in a Single Round

| | Multiple | Single | None | Improvement (Fold) |
|---|---|---|---|---|
| ECG — Poly2 | 0.234105 | 1.102187 | 2115.546985 | 3.71 |
| ECG — Poly3 | 2.160822 | 11.429962 | 16743.767084 | 4.29 |
| DRT — Poly2 | 6.838521 | 25.827835 | 1597.192878 | 2.78 |
| DRT — Poly3 | 7.234008 | 25.777343 | 1652.188852 | 2.56 |
| DRT — RBF | 6.997008 | 28.316223 | 1620.388448 | 3.05 |

*RBFs are inapplicable to intrinsic space due to infinite dimensions. Improvement is computed based on comparison between multiple and single incremental analyses

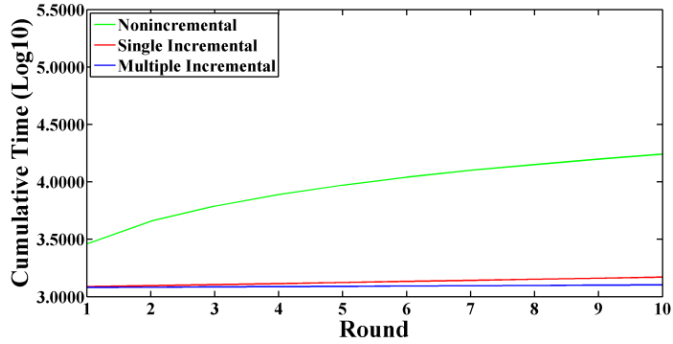

Fig. 6. KRR comparison between multiple incremental (blue), single incremental (red) and nonincremental (green) learning with the use of the DRT dataset and the RBF. The computational time (log10) was cumulative. The accuracy rates were all 90.00%.

Fig. 2–Fig. 6 display the incremental/nonincremental results, where the horizontal axis is the round, and the vertical axis represents the cumulative computational time in log10. The unit was seconds. The green curve represents the nonincremental analysis, and the red curve indicates the single incremental method. The proposed approach is shown by the blue curve. For the computational time of a single round, Table III–Table VII display the details of single rounds, and average computational time is summarized in Table VIII. Examining the result in Table VIII reveals that the proposed mechanism could improve the efficiency in intrinsic space by more than 3.71 times and the performance in empirical space by more than 2.56 times, compared with the single incremental algorithm.

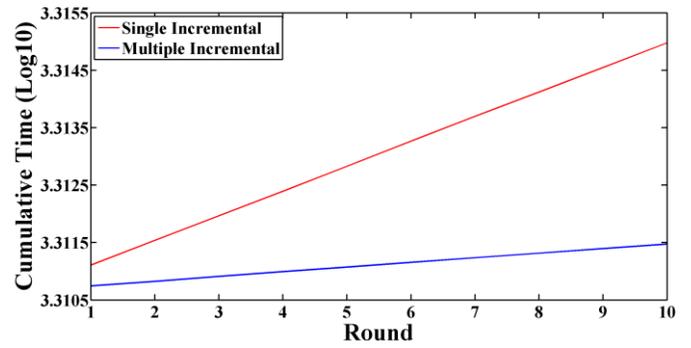

Fig. 7. KBR comparison between multiple incremental (blue) and single incremental (red) learning with the use of the ECG dataset and the poly2 kernel. The computational time (log10) was cumulative.

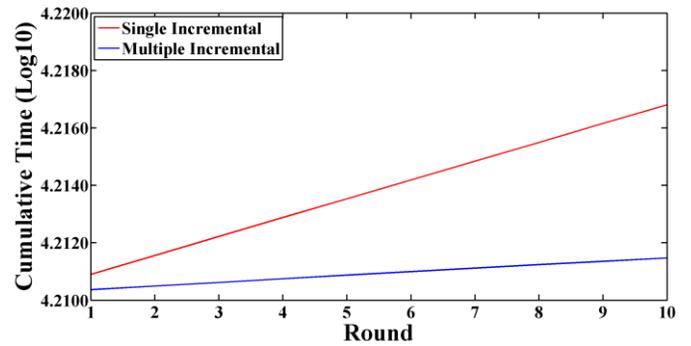

Fig. 8. KBR comparison between multiple incremental (blue) and single incremental (red) learning with the use of the ECG dataset and the poly3 kernel. The computational time (log10) was cumulative.

As for KBR, the same dataset along with the same settings was used for evaluation. The details are listed at the beginning of this section. Fig. 7–Fig. 8 show the cumulative computational time based on the proposed method and the single incremental algorithm. The detailed time is listed in Table X–Table XI. The average computational time is listed in Table XII

Table X
KBR Computational Time (log10) Based on the ECG Dataset and the Poly2 Kernel in a Single Round

| #Samples | 83226 | 83228 | 83230 | 83232 | 83234 | 83236 | 83238 | 83240 | 83242 | 83244 |
|---|---|---|---|---|---|---|---|---|---|---|
| **Multiple** | -0.433992 | -0.432390 | -0.386889 | -0.407412 | -0.425207 | -0.416759 | -0.411944 | -0.435774 | -0.419782 | -0.430701 |
| **Single** | 0.316193 | 0.308310 | 0.304919 | 0.306496 | 0.311477 | 0.316939 | 0.309808 | 0.304631 | 0.309345 | 0.308108 |

Table XI
KBR Computational Time (log10) Based on the ECG Dataset and the Poly3 Kernel in a Single Round

| #Samples | 83226 | 83228 | 83230 | 83232 | 83234 | 83236 | 83238 | 83240 | 83242 | 83244 |
|---|---|---|---|---|---|---|---|---|---|---|
| **Multiple** | 0.647582 | 0.670926 | 0.664598 | 0.687889 | 0.675405 | 0.656299 | 0.670020 | 0.650114 | 0.637448 | 0.647175 |
| **Single** | 1.385879 | 1.390621 | 1.395218 | 1.395784 | 1.387192 | 1.395385 | 1.390216 | 1.392410 | 1.401707 | 1.388583 |

Table XII
KBR Average Computational Time in a Single Round

|  | Multiple | Single | Improvement (Fold) |
|---|---|---|---|
| **ECG — Poly2** | 0.380326 | 2.040052 | 4.36 |
| **ECG — Poly3** | 4.581399 | 24.678768 | 4.39 |

*RBFs are inapplicable to intrinsic space due to infinite dimensions

## VI. Conclusion

This work presents an efficient incremental/decremental mechanism for updating the weight vector of KRR functions. The proposed mechanism combines data insertion and deletion together in the same equation, such that operations on data modifications are performed in the same round. This mechanism is conducive to improvement of computational loads, and it becomes more efficient than typical single-instance incremental analysis. Moreover, this work also presents intrinsic-space and empirical-space updates. The former is suitable for the case with $N > M$, whereas the latter fits the case when $N < M$. This study also suggests an appropriate batch size during multiple incremental/decremental analyses in intrinsic and empirical space. For intrinsic space, the mathematical model shows that the size of each batch should be smaller than the feature dimensional size after kernel mapping. Furthermore, in empirical space when decremental computation is performed, the size of the residual data should be larger than that of samples that are about to be removed. Otherwise, both situations save no computation. Finally, this study employed the developed incremental and decremental mechanism for KBR to speed up uncertainty calculation.

Open benchmark datasets, consisting of two typical datasets where $N > M$ and $M > N$, were used to evaluate the computational performance. Compared with the single incremental algorithm, the computational speed of the proposed method for KRR was enhanced by more than 3.71 times in intrinsic space and by more than 2.56 times in empirical space. For KBR, computational speed was 3.38-fold faster than the single incremental one on average. Such findings have established the effectiveness of the multiple incremental/decremental analyses.

## VII. Acknowledgements

This work was supported by 2016 Strategic Large Grants of Monash University with Medtronic Inc.